\documentclass{article}
\usepackage{graphicx} 
\usepackage{hyperref}
\usepackage{natbib} 
\usepackage{booktabs}
\usepackage{comment}
\usepackage{mdframed}
\usepackage{CJKutf8} 

\usepackage[margin=1in]{geometry} 

\usepackage{multirow}

\title{Breeze-7B Technical Report}
\author{MediaTek Research\thanks{Please cite this paper as ``MediaTek-Research (2024)". Authorship contribution details are provided at the end of this document.}}

\date{February 2024}

\begin{document}

\maketitle

\begin{abstract}
    Breeze-7B is an open-source language model based on Mistral-7B, designed to address the need for improved language comprehension and chatbot-oriented capabilities in Traditional Chinese. This technical report provides an overview of the additional pretraining, finetuning, and evaluation stages for the Breeze-7B model. The Breeze-7B family of base and chat models exhibits good performance on language comprehension and chatbot-oriented tasks, reaching the top in several benchmarks among models comparable in its complexity class. 

\end{abstract}

\section{Introduction}
This technical report presents \textit{Breeze-7B}, the latest addition to MediaTek Research's suite of language models. \textit{Breeze-7B} represents a step forward in the development of large language models (LLMs) specifically tailored for Traditional Chinese language tasks. In this report, we detail the methodologies from data curation and preparation, architectural adaptations, model training, and evaluation results of \textit{Breeze-7B}. \\ 

LLMs have perform tremendously well in language understanding task including reasoning and text summarization, resulting in them used commercially in question-and-answer platforms and customer services. Some of the most performant models are closed-source in nature, e.g.  GPT-4 \citep{openai2023gpt4}, Gemini-1.5 \citep{Google2024Gemini}, Claude. On the other hand, there are open sourced models with suitable licensing terms including Mistral \citep{jiang2023mistral}, LLaMA-2 \citep{touvron2023llama}, Falcon \citep{almazrouei2023falcon}, and BLOOM \citep{scao2022bloom}. Despite the great amount of effort in the field, typically these LLMs are optimized for English usage. Even though they are compatible with other languages, LLM behavior in non-English situations can degrade a lot. For a trivial example, the compression ratio achieved by the tokenizer can suffer \citep{sennrich-etal-2016-neural}. There is an obvious need to have open source LLMs with improved Traditional Chinese capabilities.\\

Building upon the impressive foundation laid by the \textit{Mistral-7B} model, \textit{Breeze-7B} was developed with addressing the specific challenges associated with processing and understanding Traditional Chinese text in mind. The scale of additional pretraining and finetuning is order of magnitude greater than our past attempts \citep{bloomzh2023, model7c2023}. On the set of benchmarks that we tested, \textit{Breeze-7B} performs well in Traditional Chinese tasks compared with modern models in the similar complexity class. It also maintains all the remarkable capabilities inherited from \textit{Mistral-7B}.  \\


In order to enable the community to build on this work and contribute to the development of Traditional Chinese LLMs, we are open-sourcing the model weights of \textit{Breeze-7B-Base} and \textit{Breeze-7B-Instruct}. The former is the self-supervised model additionally pretrained on Traditional Chinese corpus. The latter further finetunes on the base model to enhance instruction following and chatting abilities.



\section{Method}
\subsection{Model architecture customization}

To increase the compression rate of the Traditional Chinese corpus, we employed a modified tokenizer with additional tokens, which is a strategy also present in domain adapted LLMs \citep{liu2023chipnemo}. This change has the extra benefit of speeding up both Traditional Chinese training and inference. We use the BPE algorithm \citep{sennrich-etal-2016-neural} on a Chinese corpus to generate tokens to appended to the original tokenizer. After de-duplicating identical tokens already present in Mistral tokenizer, we extended 29873 tokens to it to arrive at 61872 total tokens. The input embedding layer and output embedding layer shapes are adjusted accordingly to accommodate the embeddings required for the new tokens. With the extended vocabulary, the compression rate on the Chinese corpus is around 2x compared to that of the Mistral model. As a result, our model enjoys a 11.1k context length on Traditional Chinese text, equivalent to around 10 pages of text, and we also observed a 2x speed up in the model training and inference speed. In addition, our long model further quadruples the context length. 



\subsection{Training}
\label{sec:training}
We trained the model on 650 gigabytes of data using a total of 7,000 H100 hours to arrive at \textit{Breeze-7B-Base-v1\_0}. Most of the training was conducted with tensor parallel and data parallel training using the Megatron-LLM library \citep{epfmgtrn}. We use a flat learning rate of 1e-6 and a batch size of 4 million tokens. The \textit{Breeze-7B-Base-v0\_1} is a earlier checkpoint of this run. \\

\begin{figure}[h!]
  \centering
  \includegraphics[width=0.5\textwidth]{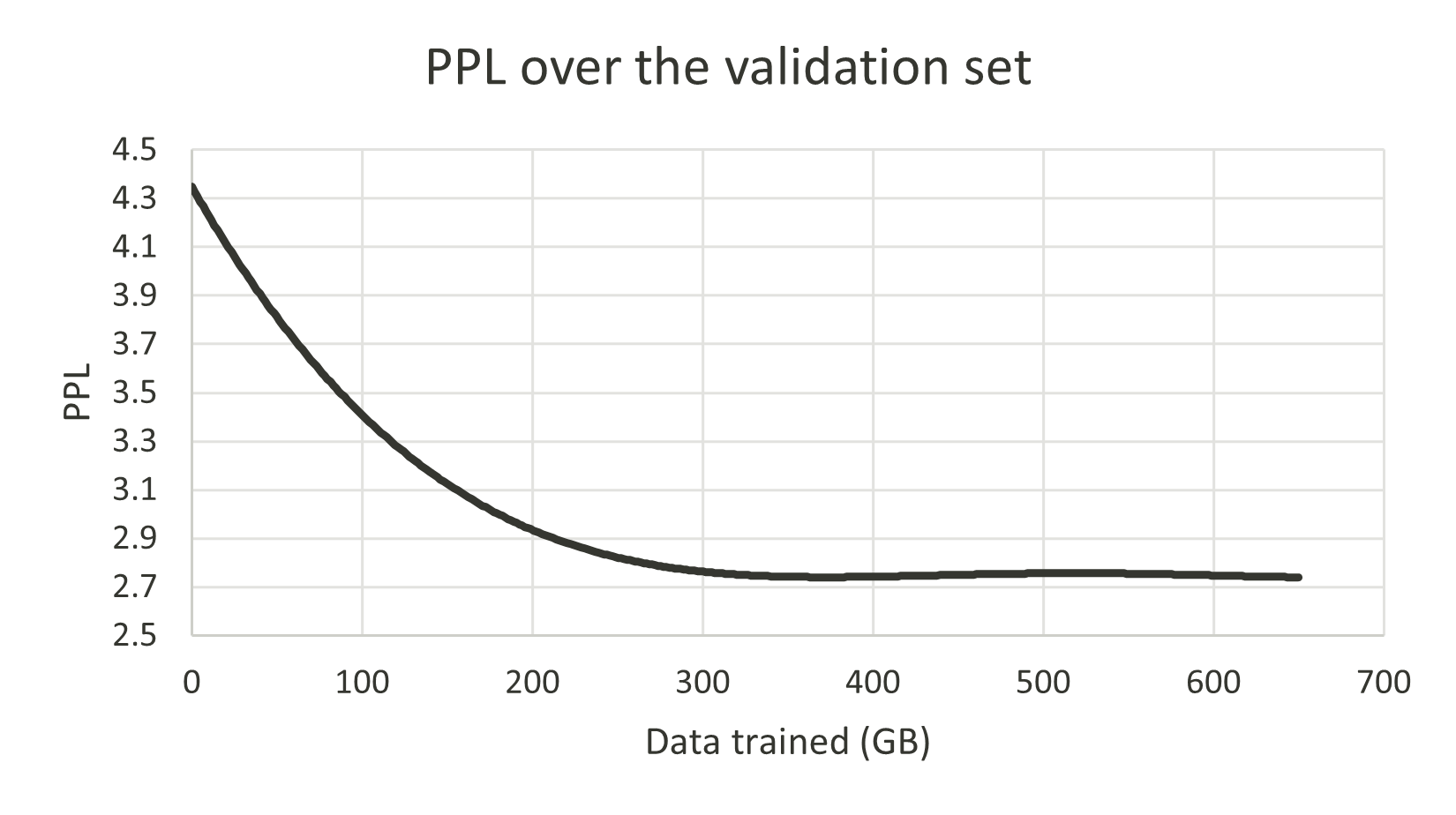}
  \caption{Perplexity (PPL) change during the additional pretraining stage of \textit{Breeze-7B}, after the vocabulary size extension. The PPL scores are calculated using our proprietary Traditional Chinese validation dataset.}
  \label{fig:training_curve}
\end{figure}

The quality of the data is of great importance of the model training, and recent works correlate data quality to textbook-like data \citep{li2023textbooks,gunasekar2023textbooks}. We found that a good validation dataset is quite important for in-training evaluation, that is, for identifying problematic checkpoints in time during the pretraining process.
In order to measure ``goodness'' of the intermediate checkpoints efficiently, we chose perplexity as a measure to align with the training objective of the model, and we selected unseen textbook-like data to highlight the potential knowledge storage in the model. \autoref{fig:training_curve} shows the perplexity changes over the training steps.

\subsection{Long Context Pretraining}
The extension from short-context to long-context models is an important field of research, as this two-step approach has significant efficiency advantages over direct long-context training without performance degradation \citep{xiong2023effective}. For this setting, different scaling methods exist \citep{chen2023extending, NTK-aware, peng2023yarn} to either induce a better inductive bias from the short-range trained positions or obtain better extrapolation properties \citep{liu2024scaling}. For our training, we incorporate a base-changing method, driven by the effectiveness and conciseness of the method on other open-sourced models \citep{jiang2023mistral,ai2024yi}. We trained the model on an additional 25 gigabytes of data consisting of Chinese, English and code data, and we upsampled long data for better long-range attention modeling. We turned off the sliding window function and set the context length to 32k, and keep all hyperparameters same as section~\ref{sec:training}

\subsection{Instruction Finetuning}
To enable the \textit{Breeze-7B} to complete question answering in a human-preferred instructional or chatting format, we fine-tuned the model with additional paired data. All data that we selected for finetuning are derived from publicly available instruction datasets, most of which are hosted on HuggingFace. For some datasets, we use our in-house filtering and transformation methods, such as deduplication and harmfulness prompt filtering, to create variants that better suit our needs. We fine-tuned the model for multiple epochs with a cosine annealing learning rate \citep{loshchilov2016sgdr}. We name the resulting model as \textit{Breeze-7B-Instruct}, with the version tag following its base model.

\section{Benchmarks}

In this section, we give a quick overview on the benchmarks we use. We considered two important aspect of language models: the ability to understand languages and solve tasks, and the ability to chat with humans.

\subsection{Language comprehension benchmarks}
We chose on TMMLU+ \citep{ikala2023eval}, DRCD \citep{Shao2018DRCDAC} and Penguins-in-a-table-TC of TC-Eval to evaluate the models. TMMLU+ is a benchmark derived from TMMLU \citep{model7c2023}, curated from examinations in Taiwan. It consists of 67 subjects spanning multiple disciplines, from vocational to academic fields, and covers elementary to professional proficiency levels. It is further categorized into 4 subcategories similar to MMLU: STEM, humanities, social sciences and others. DRCD is an extractive question answering benchmark based on Wikipedia. Penguins-in-a-table-TC is a multiple choice benchmark focusing on table understanding abilities, including table operations such as insertion and sorting.

\subsection{Chatbot-oriented benchmarks}
We chose MT-Bench-tw and MT-Bench \citep{zheng2024judging} to reflect chatting abilities of the language models. MT-Bench-tw is multi-turn dialogue benchmark translated from MT-bench, with slight adjustments in some terms to reflect the language specific task instructions that do not translate well to Traditional Chinese, for example, "Start each sentence with the letter A". The multi-turn benchmark consists of two turns. In the first turn, versatile questions covering multiple topics of interest are used to test the chatting abilities of the models. In the second turn, further instructions would be given to test the models' abilities to follow verbose instructions and attend to inputs and outputs of the previous round. Due to its free-form answering nature, we follow the convention of scoring with GPT-4 on a scale from 1 to 10. We provide the exact prompt for scoring in Appendix~\ref{sec:apx-chat}.

\subsection{Long-context benchmarks}
To understand the models' ability to attend to long context, we collected benchmarks that have inputs spanning over 10k tokens. At this scale, the benchmarks serve as a precursor to the performance on document-level tasks, as well as the commercially popular RAG (Retrieval Augmented Generation) use case. We first benchmark the model on the passkey retrieval task following previous work, and then on DRCD-long that is synthetically created from the Chinese benchmark DRCD. Passkey retrieval is a synthetic task to test whether a model could retrieve a 5-letter key within a long paragraph of irrelevant text. In addition to the variation of total sequence length, we additionally control the key position following observations that the depth of the key can significantly affect performance. DRCD-long extends the DRCD context to up to 64k Chinese characters with irrelevant articles, simulating inaccuracies in an RAG system. The task objective and scoring criterion remain the same as the original DRCD. We leave the curation details of this benchmark in Appendix~\ref{sec:apx-long-bench}.
\section{Results}

In this section, we reported the evaluated results of our model, and contrasted our results with those of several notable contemporary models. 
Since some of the models have multiple variants, we select model variants that maximize performance. For the language comprehension benchmarks, we use the instruct variants whenever possible; otherwise, we resort to the chat variants. We only use the Chat variants for the chatbot-oriented benchmarks. Additionally, to understand the effectiveness of the pretraining of different models, we also evaluate the language comprehension benchmarks on the base model. To increase robustness, we used a few-shot template to demonstrate the task format.

\subsection{Models evaluated}

We evaluate the following models over the language comprehension benchmarks. We chose the models based on two main factors: 1. Bilingual (Chinese/English) Abilities 2. Similar size to Breeze-7B
\begin{table}[h]
\centering
\begin{tabular}{|c|c|}
\hline
\textbf{Base Model}       & \textbf{Instruct/Chat Model}                \\ \hline
Breeze-7B-Base            & Breeze-7B-Instruct              \\ \hline
Yi-6B                     & Yi-6B-Chat                           \\ \hline
Qwen1.5-7B                   & Qwen1.5-7B-Chat      \\ \hline
(unreleased at the moment of the writing of this paper) & Taiwan-LLM-13B-v2.0-Chat             \\ \hline
(unreleased at the moment of the writing of this paper) & Taiwan-LLM-7B-v2.1-Chat             \\ \hline

\end{tabular}
\caption{Evaluate Base and Chat/Instruct Models}
\label{table:base_chat_models}
\end{table}


Also, we listed the benchmark scores of GPT-3.5-Turbo (1106), which represents one of the most widely used high-quality cloud language model API services, for reference.

\subsection{Results}

\subsubsection{Traditional Chinese Language Comprehension benchmarks}
Table~\ref{table:chat-comp-zh} and Table~\ref{table:base-comp} summarizes the comprehension benchmark results on the chat model and the base model respectively. Since the base model may not reliably identify the format of multiple-choice questions, we have employed a 5-shot setting to enhance the stability and fairness of the comparison, using five examples from the development set as the shots. However, due to context limitations, the DRCD was reduced to a 3-shot setting. 

From the results, we can draw three conclusions. First, the \textit{Breeze-7B-Instruct} model performs well on TMMLU+ compared to other models of the same size, while outperforming other models in the Penguins-in-a-table-TC dataset. Second, \textit{Breeze-7B-Instruct} demonstrates superior or comparable performance to the \textit{Breeze-7B-Base} model, indicating that our finetuning procedure effectively retains the knowledge in the base model.  Finally, when comparing the performance of \textit{Breeze} and \textit{Mistral} on the MMLU benchmark, we conclude that our training procedure, along with adjustments to the model architecture, only results in minimal degradation of English performance.
\begin{table}[h!]
\centering
\small
\begin{tabular}{|c|c|c|c|c|c|c|c|}
\hline
\multirow{2}{*}{\textbf{Model}} & \multicolumn{5}{c|}{\textbf{TMMLU+(ACC)}} & \textbf{Table}& \textbf{MMLU} \\
\cline{2-6}
 & \textbf{STEM} & \textbf{Social Science} & \textbf{Humanities} & \textbf{Other} & \textbf{$\uparrow$ AVG} &(ACC)&(ACC) \\
\hline
 Yi-6B-Chat & 37.80 & 51.74 & 45.36 & 44.25 & 44.79 & 25.69 & 59.45\\
 GPT-3.5-Turbo & 41.58 & 48.52 & 40.96 & 43.18 & 43.56 & 45.14 & 67.09 \\
 \textbf{Breeze-7B-Instruct-v1\_0} & 36.46 &48.37 & 45.11 & 40.75 & 42.67 & 39.58 & 59.12\\
 Qwen1.5-7B-Chat &  41.48& 51.66 & 44.05 & 45.40 &45.64  & 34.72  & 59.54\\
 Taiwan-LLM-13B-v2.0-chat & 27.74 & 33.69 & 27.03 & 29.43 & 29.47 & 23.61& 50.50 \\
 Taiwan-LLM-7B-v2.1-chat & 25.58 & 31.76 & 27.36 & 27.61 & 28.08 & 31.25 & 42.72\\
\hline
\end{tabular}
\caption{Benchmark results of finetuned models on TMMLU+ (0 shot), Table (0 shot), and MMLU (0 shot)}
\label{table:chat-comp-zh}
\end{table}

\begin{table}[h!]
\centering
\begin{tabular}{|c|c|c|c|c|}
\hline
\textbf{Model}  & \textbf{TMMLU+ (ACC)} & \textbf{DRCD (EM)} & \textbf{Table (ACC)} & \textbf{MMLU (ACC)} \\
\hline

Yi-6B  & 49.63 & 76.61 & 34.72 & 65.35 \\

Qwen1.5-7B & 46.28 & 74.41 & 30.56 & 60.53 \\
\textbf{Breeze-7B-Base-v1\_0}  & 40.72 & 80.61 & 31.99 & 58.65 \\

Mistral-7B-v0.1 & 36.93 & 79.27 & 27.78 & 64.89 \\
\hline
\end{tabular}
\caption{Benchmark results of pretrained models on TMMLU+ (5 shot), DRCD (3 shot),  Table (5 shot), and MMLU (5 shot).}
\label{table:base-comp}
\end{table}
\subsubsection{Traditional Chinese Chatbot-oriented benchmarks}
Table~\ref{table:chat-chat}  summarizes the benchmark scores for chatbot-oriented models. Overall, \textit{Breeze-7B-Instruct} performs well on both the Chinese and English versions of the benchmark, with \textit{GPT-4} serving as the judge. It trails only behind \textit{Qwen1.5-7B}. We present the subcategory scores for the 10 total categories in Appendix~\ref{sec:apx-chat}

\begin{table}[h!]
\centering
\begin{tabular}{|c|c|c|}
\hline
\textbf{Models} & \textbf{MT-Bench-tw (Score)} & \textbf{MT-Bench (Score)} \\
\hline
GPT-3.5-Turbo &7.1& 7.9 \\
\hline
\textbf{Breeze-7B-Instruct-v1\_0} &6.0& 7.4 \\
\hline
Qwen1.5-7B-Chat & 6.4 & 7.6 \\
\hline
Yi-6B-Chat &5.0& 6.0 \\
\hline
Taiwan-LLM-13B-v2.0-chat &5.0& N/A* \\
\hline
Taiwan-LLM-7B-v2.1-chat &4.2& N/A* \\
\hline
\end{tabular}
\caption{Benchmark results of finetuned models on MT-Bench score. Results marked in asterisk should not be interpreted as the models true capabilities as the model often answered in Chinese in this English benchmark, causing scoring difficulties. }
\label{table:chat-chat}
\end{table}

\subsubsection{Long Context Benchmarks}
We compare \textit{Breeze-7B-32k-Base} with \textit{Breeze-7B-Base} to show the effect after long context pre-training. As our focus of this paper is on enabling the long-context abilities of Breeze-7B-Base, we leave comparisons to other models as future work. 

\begin{figure}[h!]
  \centering
  \includegraphics[width=1\textwidth]{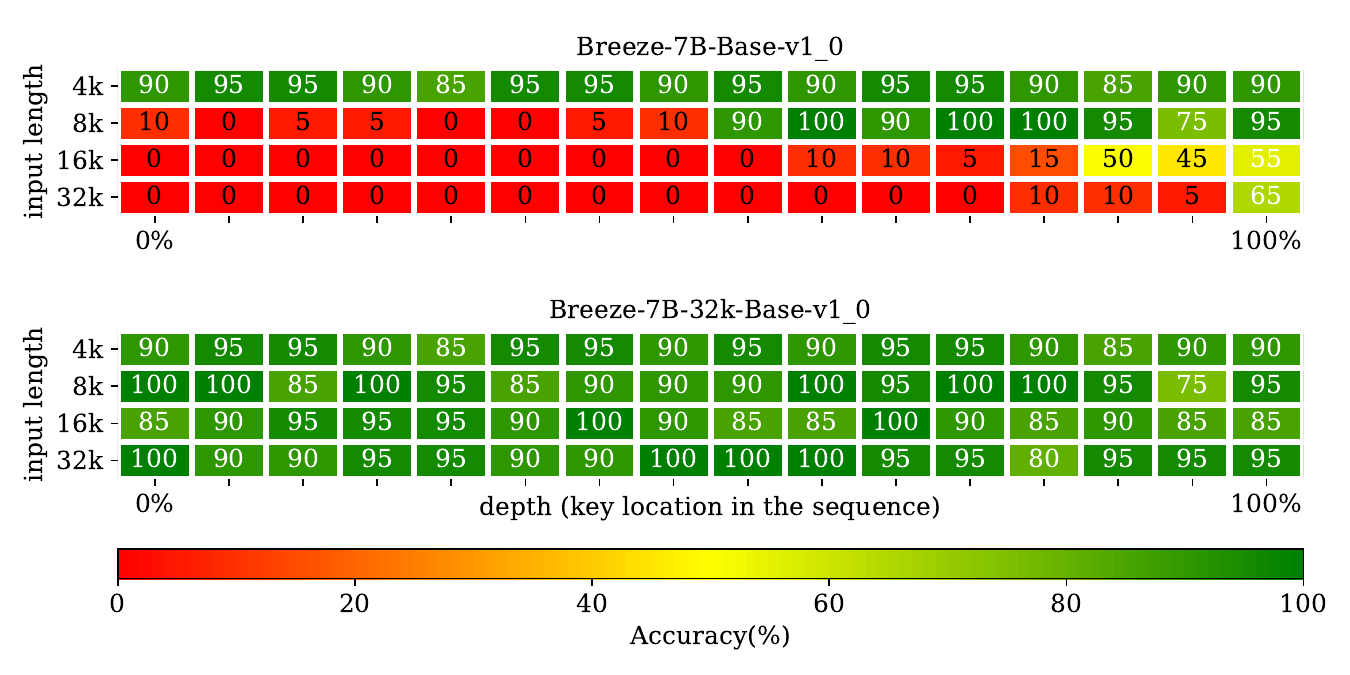}
  \caption{Passkey Retrieval results of \textit{Breeze-7B-Base} and \textit{Breeze-7B-32k-Base}. The y-axis denotes the input sequence length, while the x-axis denotes the depth of the key position in the example. Each length-depth combination is trialed 20 times and the accuracy is color-coded with the colormap at the bottom.} 
  \label{fig:passkey}
\end{figure}

For the passkey retrieval task, we evenly split the key positions into 16 bins and tested 20 examples per bin. Accuracy is reported for each length-depth combination, and we visualize the results in Figure~\ref{fig:passkey}. The non-extended model can only consistently retrieve keys where they are located within a 4k token distance from the question, which is positioned at the end of the sequence. This phenomenon can be attributed to the sliding window size of non-extended model (4096), which limits the direct attention span, that is, information beyond this would have to rely on "information forward passing"\citep{jiang2023mistral} to be attended to. As expected, after long-context training, \textit{Breeze-7B-32k-Base} performs consistently on the task up to 32k. An analysis of the incorrect examples reveals that almost all errors are copying errors, such as mistaking lowercase letters as uppercase, which indicates that the model has near-perfect performance in attending to the correct position. 

For Long-DRCD, We observed in Table~\ref{table:drcd-long} that \textit{Breeze-7B-32k-Base} has a large boost in performance in 3-shot DRCD compared to \textit{Breeze-7B-Base}, and the model tested on DRCD-32k retains 77\% of the DRCD performance. The ability to attend to text very far away benefits twofold: finding information that occurs in the front of the context and understanding the extractive setting of the task demonstrated by few-shot examples in the very front. We also include DRCD-64k which exceeds our training context length. At this length, the degradation is observable for both models, showing that our training method does not extrapolate well to longer sequences. 
\begin{table}[h!]
\centering
\begin{tabular}{|c|c|c|c|c|}
\hline
\textbf{Model/Perforamce(EM)}  & \textbf{DRCD} & \textbf{DRCD-16k} & \textbf{DRCD-32k} & \textbf{DRCD-64k} \\
\hline

\textbf{Breeze-7B-32k-Base-v1\_0} & 79.73 & 69.68 & 61.55 & 25.82 \\
\textbf{Breeze-7B-Base-v1\_0}  & 80.61 & 21.79 & 15.29 & 12.63 \\

\hline
\end{tabular}
\caption{Benchmark results of pretrained models on DRCD with different context length (3-shot).}
\label{table:drcd-long}
\end{table}

\section{Conclusion}

In this study, we have introduced \textit{Breeze-7B}, a Traditional Chinese enhanced language model based on \textit{Mistral-7B}. These base and instruct variants of the model have shown great performance on language comprehension and chatting abilities, closing the gap between open sourced models and proprietary models such as GPT-3.5.  \textit{Breeze-7B} stands as a testament to MediaTek Research's commitment to the realm of artificial intelligence and natural language processing, and we plan to keep pushing the boundaries of what is possible with language models.

\appendix

\section*{Appendix}



\section{Chat Benchmark Details}
Our GPT-4 judge incorporates the following prompt for MT-Bench-tw, where words in curly brackets will be replaced with real questions in the benchmark and answers of the language model.
\begin{mdframed}
$[$Instruction$]$\\
Please act as an impartial judge and evaluate the quality of the response provided by an AI assistant to the user question displayed below. Your evaluation should consider factors such as the helpfulness, relevance, accuracy, depth, creativity, and level of detail of the response. Responses in Traditional Chinese are expected and are more favourable than Simplified Chinese, English, and other languages. Begin your evaluation by providing a short explanation. Be as objective as possible. After providing your explanation, you must rate the response on a scale of 1 to 10 by strictly following this format: [[rating]], for example: Rating: [[5]].\\\\
$[$Question$]$\\\\
\{question\}\\\\
$[$The Start of Assistant's Answer$]$\\\\
\{answer\}\\\\
$[$The End of Assistant's Answer$]$
\end{mdframed}
Table ~\ref{table:chat-chatverbose} shows  scores of the finetuned model on all 8 subcategories of the MT-Bench-tw benchmark. \textit{Breeze-7B-Instruct} performs competitively with other models overall in the 7B size range, but it trails behind other models to greater extent on STEM and Writing subtasks. Enhancements in these areas is an exciting direction of future work.
\label{sec:apx-chat}
\begin{table}[h]
\tiny
\centering
\begin{tabular}{|c|c|c|c|c|c|c|c|c|c|}
\hline
\textbf{Model} & \textbf{STEM} & \textbf{Extraction} & \textbf{Reasoning} & \textbf{Math} & \textbf{Coding} & \textbf{Roleplay} & \textbf{Writing} & \textbf{Humanities} & \textbf{$\uparrow$ AVG} \\ \hline
GPT-3.5-Turbo & 7.8 & 6.1 & 5.1 & 6.4 & 6.2 & 8.7 & 7.4 & 9.3 & 7.1 \\ 
\textbf{Breeze-7B-Instruct-v1\_0} & 7.8 & 5.2 & 4.2 & 4.2 & 4.1 & 7.6 & 5.9 & 9.1 & 6.0 \\ 
Qwen1.5-7B-Chat & 9.0 & 5.6 & 4.7 & 2.8 & 3.7 & 8.0 & 8.0 & 9.4 & 6.4 \\
Yi-6B-Chat & 7.3 & 2.7 & 3.1 & 3.3 & 2.3 & 7.2 & 5.2 & 8.8 & 5.0 \\ 
Taiwan-LLM-13B-v2.0-chat & 6.1 & 3.4 & 4.1 & 2.3 & 3.1 & 7.4 & 6.6 & 6.8 & 5.0 \\ 
Taiwan-LLM-7B-v2.1-chat & 5.2 & 2.6 & 2.3 & 1.2 & 3.4 & 6.6 & 5.7 & 6.8 & 4.2 \\ \hline
\end{tabular}
\caption{Verbose benchmark results of finetuned models on MT-Bench-tw}
\label{table:chat-chatverbose}
\end{table}

\section{Long Context Benchmark Details}
\label{sec:apx-long-bench}
As seen in the passkey retrieval results, the position of the relevant information can greatly influence the performance. To ensure that the correct context span is even distributed with various depth across samples, our curation method of the DRCD-long tasks are as follow: For questions ${a,b,c,d,e...}$ that require context ${A,B,C,D,E...}$, we concatenate the contexts until they reach the target length to arrive at ${ABCDE, FGHIJ, ...}$ as the contexts. The questions are automatically paired with the extended context. We have released DRCD-long at our \href{https://huggingface.co/datasets/MediaTek-Research/TCEval-v2}{huggingface page}
\bibliographystyle{plainnat}
\bibliography{reference}

\section*{Authorship and Acknowledgements}
\begin{CJK*}{UTF8}{bsmi}
This work is a collaborative effort by the following members:  (in alphabetical order): Chan-Jan Hsu 許湛然, Chang-Le Liu 劉昶樂, Feng-Ting Liao 廖峰挺, Po-Chun Hsu 許博竣, Yi-Chang Chen 陳宜昌, and the supervisor Da-Shan Shiu 許大山.
\end{CJK*}

We thank Nvidia-Taiwan for providing computational and storage resources.
\end{document}